\documentclass[pdflatex,sn-vancouver,Numbered]{sn-jnl}

\usepackage{graphicx}%
\usepackage{multirow}%
\usepackage{amsmath,amssymb,amsfonts}%
\usepackage{amsthm}%
\usepackage{mathrsfs}%
\usepackage[title]{appendix}%
\usepackage{xcolor}%
\usepackage{textcomp}%
\usepackage{manyfoot}%
\usepackage{booktabs}%
\usepackage{algorithm}%
\usepackage{algorithmicx}%
\usepackage{algpseudocode}%
\usepackage{listings}%

\begin{document}

\title[Distinguishing performance gains from learning]{Distinguishing performance gains from learning when using generative AI}

\author*[1]{\fnm{Lixiang} \sur{Yan}}\email{lixiang.yan@monash.edu}
\author[2]{\fnm{Samuel} \sur{Greiff}}
\author[3]{\fnm{Jason M.} \sur{Lodge}}
\author[1]{\fnm{Dragan} \sur{Gašević}}

\affil*[1]{\orgdiv{Faculty of Information Technology}, \orgname{Monash University}, \orgaddress{\city{Clayton}, \state{Victoria}, \country{Australia}}}
\affil[2]{\orgdiv{School of Social Sciences and Technology}, \orgname{Technical University of Munich}, \orgaddress{\city{Munich}, \country{Germany}}}
\affil[3]{\orgdiv{School of Education}, \orgname{The University of Queensland}, \orgaddress{\city{Brisbane}, \state{Queensland}, \country{Australia}}}

\maketitle

\begin{center}
\textbf{Accepted manuscript / postprint version}
\end{center}

\noindent This is the accepted manuscript version of the following article:

\medskip

\noindent Yan, L., Greiff, S., Lodge, J. M., \& Gašević, D. (2025). Distinguishing performance gains from learning when using generative AI. \textit{Nature Reviews Psychology, 4}, 435--436. https://doi.org/10.1038/s44159-025-00467-5

\medskip

\noindent The final authenticated version is available online at: 

\noindent https://www.nature.com/articles/s44159-025-00467-5

\medskip

\noindent © Springer Nature Limited 2025. This accepted manuscript is made available in accordance with the Springer Nature self-archiving policy. This manuscript may not be redistributed or reused for commercial purposes and may not be released under a Creative Commons licence.

\section*{Standfirst}

Generative artificial intelligence (AI) is increasingly being integrated into education, where it can boost learners’ performance. However, these uses do not promote the deep cognitive and metacognitive processing that are required for high-quality learning.

\section*{Main}

The rapid adoption of generative artificial intelligence (AI) tools such as ChatGPT is transforming how learners engage with information, especially in tertiary education. Celebrated for providing instant assistance and personalized feedback, these tools promise to enhance performance across tasks such as academic writing, computer programming and information research\citep{deng2025,yan2024}. Benefits such as improved task performance and reduced cognitive load have been reported for tertiary students working on writing and information search tasks\citep{stadler2024}. However, beneath these immediate gains lies a paradox: although generative AI can boost learners’ performance on tasks, it risks undermining the cognitive and metacognitive processes that are essential for durable learning\citep{stadler2024,fan2024}.

There is a fundamental distinction between performance and learning\citep{soderstrom2015}. Performance refers to observable behaviours during task execution, often influenced by external factors or supports. By contrast, learning involves enduring changes in knowledge or behaviour that result from experience; learning is characterized by the ability to independently retain and transfer skills\citep{soderstrom2015}. It is well established that high performance does not always imply learning, as performance metrics are often transient and contextually dependent\citep{soderstrom2015}. For example, a student might achieve high performance on a test after massed practice (‘cramming’) the night before, yet fail to retain or apply the knowledge later, indicating performance without lasting learning\citep{soderstrom2015}.

This fundamental distinction is underexamined in emerging research on generative AI in education. For instance, a meta-analysis of 69 studies aimed to investigate the impact of ChatGPT and other generative AI tools on “student learning” in K-12 and tertiary education. Yet, the meta-analysis reported that these technologies improve student “academic performance”, with an effect of (Hedge’s) $g = 0.7$, reflecting the pervasive conflation of these concepts\citep{deng2025}. Although the reported effect size is large, it probably reflects immediate task success rather than learning. Another study identified significant boosts to undergraduate student performance in providing quality peer feedback when using generative AI-based assistance\citep{darvishi2024}. However, these performance improvements diminished once the generative AI assistance was removed, indicating that immediate gains in performance might not reflect sustained improvements in student learning.

\section*{Cognitive load, metacognition and self-efficacy}

Clarifying the distinction between performance and learning requires examination of how generative AI influences the key psychological processes involved in durable learning. Specifically, balancing cognitive load, metacognition and self-efficacy are critical to achieving deeper, lasting learning\citep{soderstrom2015,sweller2011,ryan2020}. Investigating these constructs can help to reveal whether immediate performance gains from generative AI translate into sustained improvements in knowledge and skills.

Generative AI tools can in some cases reduce cognitive load, particularly for tasks that require complex information processing. For instance, students who used ChatGPT to research socio-scientific issues reported significantly lower cognitive load compared with those who used more traditional research tools such as internet search engines\citep{stadler2024}. From the perspective of cognitive load theory\citep{sweller2011}, offloading domain-specific tasks to generative AI can free up learners’ working memory resources. However, reduced cognitive load does not necessarily lead to deeper cognitive engagement with the learning material. Indeed, excessively offloading cognitive tasks to generative AI risks reducing active cognitive engagement, which is crucial for deep encoding, effective retention and meaningful transfer. This risk is supported by evidence showing that undergraduate students who relied heavily on ChatGPT for information gathering exhibited weaker argumentation and reasoning skills compared with those who used traditional, cognitively demanding research methods\citep{stadler2024}.

Generative AI might also disrupt metacognition. A randomized controlled study indicated that frequent use of generative AI for argumentative essay writing tasks can foster “metacognitive laziness” in undergraduate students\citep{fan2024}. That is, learners offloaded critical evaluative tasks, such as reviewing task criteria, assessing written drafts against rubrics and reflecting critically on learning materials, to the tool rather than actively reflecting on their work. These evaluative tasks are crucial because they enable learners to independently judge the quality of their work, identify areas that need improvement and actively regulate their own learning process. This offloading undermines key self-regulation skills by reducing learners’ active involvement in these metacognitive activities. Psychological theories of over-reliance suggest that external tools can displace internal cognitive effort, reducing learners’ ability to independently retain and transfer knowledge\citep{zhai2024}.

Offloading metacognitive activities to generative AI can also stunt intrinsic motivation, as the learner’s role in task completion diminishes\citep{fan2024}. From the perspective of self-determination theory, intrinsic motivation depends on learners’ autonomy and active engagement\citep{ryan2020}. Delegating key metacognitive tasks, such as planning, monitoring and evaluating, to generative AI might reduce learners’ active participation and sense of control, thereby decreasing intrinsic motivation. Along these lines, undergraduate learners who interacted frequently with generative-AI-powered chatbots for acquiring academic domain-specific knowledge reported lower learning autonomy relative to students who had infrequent chatbot interactions\citep{zhai2024}.

Generative AI’s paradoxical effects on self-efficacy further complicate its role in human learning. For instance, a study reported that frequent generative AI use among university students enhanced learners’ perceived confidence and efficiency during academic tasks but simultaneously increased technological dependence\citep{zhang2025}. This technological dependence can have long-term repercussions, eroding learners’ ability to engage in independent learning and reducing their resilience in the face of new challenges\citep{soderstrom2015}.

\section*{Targeting learning in research and practice}

Harnessing the potential of generative AI requires us to rethink research design and interpretation. To address the conceptual and methodological limitations identified above, we propose a research agenda that is informed by cognitive psychology. Clarifying how generative AI tools influence individuals’ encoding, consolidation and retrieval processes is essential for determining whether these technologies enable learning or merely confer short-term performance advantages. By illuminating the underlying cognitive mechanisms, researchers can better evaluate to what extent generative AI-based educational assistance promotes deep engagement, transfer of knowledge and durable skill development, rather than simply facilitating task completion.

First and foremost, researchers should seek to differentiate learning from performance. To do so, researchers should use process-oriented assessments such as retention and transfer tests to distinguish genuine learning from mere task performance. Examination of testing effects and use of delayed recall tasks can be used to evaluate durable learning in AI-supported contexts\citep{soderstrom2015}. Second, researchers should investigate how generative AI influences encoding, consolidation and retrieval processes. For instance, the relationship between cognitive load and deep encoding remains to be clarified. Paradigms from cognitive load theory can be used to explore these effects. Finally, and in conjunction with the recommendations above, researchers should study the longitudinal effects of educational generative AI tools. The long-term effects of AI tools on knowledge retention and skill development and the interaction between AI-assisted learning and prior knowledge remain to be examined.

Educators and designers of educational generative AI tools can also take steps to ensure that generative AI-supported tasks promote learning rather than just performance gains. First, they can promote active metacognition by encouraging learners to critically evaluate AI-generated outputs rather than accepting them passively. Second, educators and practitioners should balance efficiency with autonomy. Generative AI should be used to support, not replace, independent cognitive effort. Furthermore, feedback from generative AI chatbots can be blended with opportunities for unaided problem-solving to foster students’ autonomy alongside task success.

\section*{Outlook}

Although generative AI tools such as ChatGPT can enhance learners’ performance and reduce cognitive load, they might disrupt the deeper processes of encoding, retention and independent problem-solving that are necessary for durable learning. By prioritizing process-based assessments, clarifying how generative AI interacts with cognitive functions and examining long-term effects, researchers can develop a more nuanced understanding of human--AI interactions. Ultimately, generative AI must serve as an augmentation for human learning, not a replacement, and we must ensure that learners develop knowledge and skills that extend beyond the confines of generative AI-assisted tasks.

\section*{Competing interests}

The authors declare no competing interests.

\end{document}